\documentclass{article}
\usepackage{spconf,amsmath,epsfig}
\usepackage{graphicx}
\usepackage{subcaption}

\title{UNSUPERVISED STEREO MATCHING NETWORK FOR VHR REMOTE SENSING IMAGES BASED ON ERROR PREDICTION}
\name{Liting Jiang$^{1,2,3}$, Yuming Xiang$^{1,2,3}$, Feng Wang$^{1,2,3}$, and Hongjian You$^{1,2,3}$\vspace{-9pt}}

\address{1.Aerospace Information Research Institute, Chinese Academy of Sciences\\	2.Key Laboratory of Technology in Geo-spatial Information Processing and Application System,\\ Chinese Academy of Sciences, Beijing, 100190, China\\3.School of Electronic, Electrical and Communication Engineering, \\University of Chinese Academy of Sciences, Beijing, 101408, China\vspace{-16pt}}

\begin{document}
%
\maketitle
\vspace{-27pt}
\vspace{-27pt}
\begin{abstract}
\vspace{-5pt}
Stereo matching in remote sensing has recently garnered increased attention, primarily focusing on supervised learning. However, datasets with ground truth generated by expensive airbone Lidar exhibit limited quantity and diversity, constraining the effectiveness of supervised networks. In contrast, unsupervised learning methods can leverage the increasing availability of very-high-resolution (VHR) remote sensing images, offering considerable potential in the realm of stereo matching. Motivated by this intuition, we propose a novel unsupervised stereo matching network for VHR remote sensing images. A light-weight module to bridge confidence with predicted error is introduced to refine the core model. Robust unsupervised losses are formulated to enhance network convergence. The experimental results on US3D and WHU-Stereo datasets demonstrate that the proposed network achieves superior accuracy compared to other unsupervised networks and exhibits better generalization capabilities than supervised models. Our code will be available at https://github.com/Elenairene/CBEM. 
\end{abstract}
\vspace{-5pt}
\begin{keywords}
\vspace{-3pt}
Stereo matching, unsupervised learning, uncertainty, disparity estimation
\end{keywords}
\vspace{-7pt}
\section{INTRODUCTION}
\label{sec:intro}
\vspace{-7pt}
As Earth observation and computer vision technologies rapidly advance, 3D reconstruction has gained increasing attention in remote sensing. Stereo matching, a crucial step in 3D reconstruction, generates dense pixel-by-pixel matches, yielding disparity maps that enable height information extraction. The application of 3D reconstruction in remote sensing offers a novel perspective for observation and analysis of the Earth's surface, emerging as a prominent research topic.

Traditional stereo matching methods such as Semi-Global-Matching (SGM) algorithm~\cite{SGM} has been widely used. But deep learning methods have achieved better performance recently and are experiencing a significant development. GC-Net~\cite{gcnet} first adopted 3D convolutions to construct cost volume. PSMNet~\cite{psmnet} introduced spatial pyramid pooling to enlarge the receptive fields and employed stacked hourglass module. Current state-of-the-art methods mostly use multi-feature 4D cost volume which burdens memory a lot such as GANet~\cite{GANet} and CSPN~\cite{CSPN}. To alleviate this problem, CFNet~\cite{cfnet} employs uncertainty estimation to adjust disparity search range which reduces memory consumption. Bidir-EPNet~\cite{epnet} is designed as a specialized pyramid stereo matching network for VHR images. While datasets commonly used in supervised learning methods suffer from lack of diversity, unsupervised methods are unconstrained by availability of ground truth. UnFlow~\cite{unflow} proposed efficient unsupervised loss to train FlowNet without ground truth. PASM~\cite{pasm} applied attention mechanism to stereo matching task. Some unsupervised methods~\cite{transportation} have been applied in automatic driving and have promising future prospects. T. Igeta.~\cite{igeta} designed an unsupervised stereo matching method for VHR image, validating the feasibility of designing unsupervised methods for VHR images.

Inspired by above previous work, we study unsupervised stereo matching method for VHR image and one potential approach is to generate metric that quantifies disparity error in the absence of disparity ground truth. Multiple methods~\cite{cfnet}~\cite{bidisemi} utilized confidence estimation module to help assess reliability of disparity or generating search range, but those estimations are not correlated to true error. L.Chen~\cite{learning} successfully combined disparity and uncertainty estimation, achieving improvements in both tasks. H.Shi~\cite{bidisemi} employed a simple module to generate a binary confidence score for a semi-supervised task. Designing a suitable intermediary metric is essential for refining disparity without ground truth. 

In this work, we propose a novel unsupervised stereo matching network based on error prediction for VHR remote sensing images. To overcome the limitation of ground truth, we fit the confidence-aware cascade network (CACN)~\cite{CACN} to unsupervised learning manner. To bridge network confidence with disparity errors, we design a light-weight confidence based error prediction module (CBEM) and demonstrate its cross-domain ability. Leveraging the strong coarse-to-fine ability of multi-scale cascaded core model and error prediction ability of CBEM, our method outperforms the comparison methods in terms of average endpoint error (EPE) and the fraction of erroneous pixels (D1).
\vspace{-10pt}
\begin{figure*}[!t]
\vspace{-20pt}
\centering	\includegraphics[width=2.0\columnwidth]{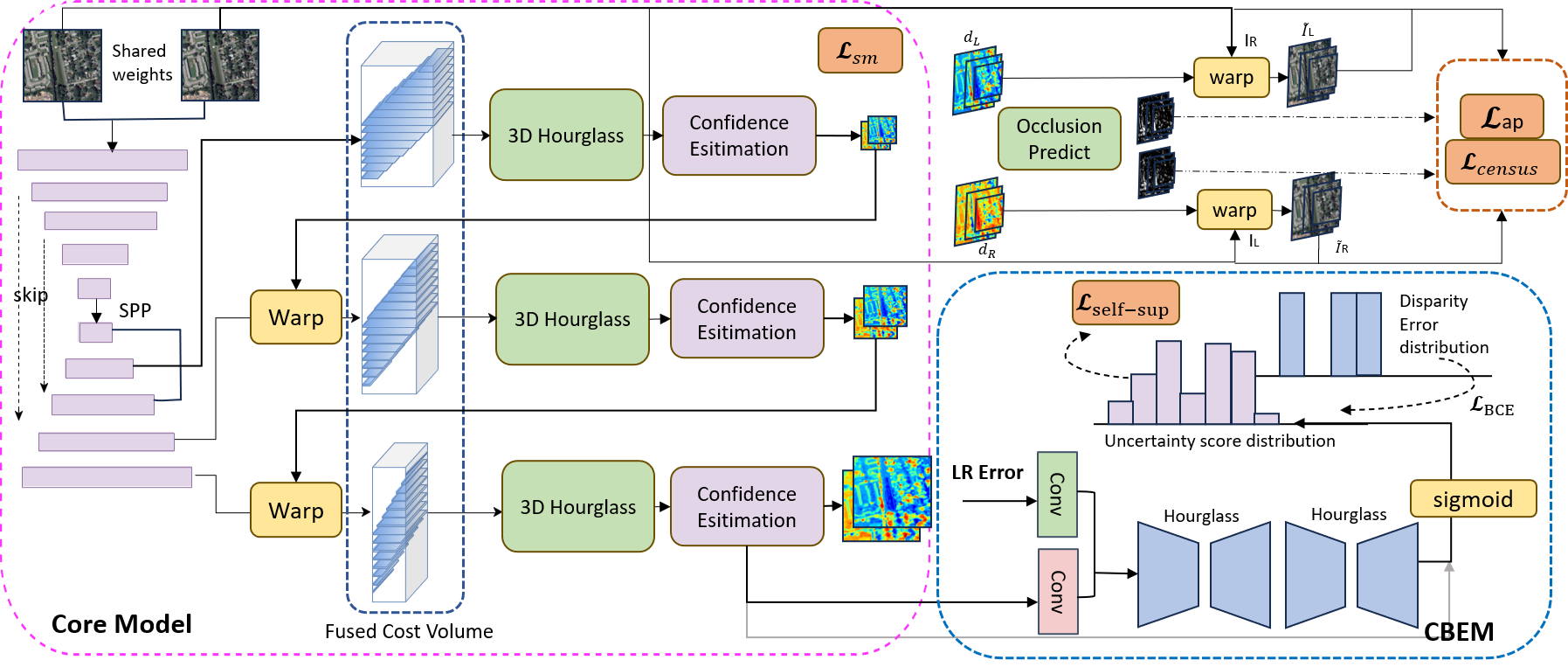}
\caption{The framework of proposed method.}
\label{fig:frame}
\vspace{-17pt}
\end{figure*}
\section{METHODOLOGY}
\label{sec:format}
\vspace{-10pt}
\subsection{The architecture of the proposed network}
\vspace{-7pt}
The complete framework of the proposed network is shown in Fig.~\ref{fig:frame}. The model proposed consists of the core model and confidence based error prediction module(CBEM), while core model is a relatively independent module that predicts disparity following unsupervised learning routine, CBEM predicts uncertainty scores closely associated with disparity errors using confidence outputs of core model. The training process follows a three-step routine: initial training of the core model, subsequent training of the CBEM with the core model fixed, and finally fine-tuning of the core model with CBEM fixed. The last two steps require only a few epochs. 

The core model is primarily constructed based on CACN ~\cite{CACN}. Below is a brief overview of the structure.  

For the feature extraction module, a multi-scale UNet-like network is employed which shares the same weights for both left and right images. Then a combination of concatenated and group-wise correlation modules is utilized to construct cost volume. The initial disparity is generated by softmax operation on cost volume with a range of $[-d_{max}^i,d_{max}^i]$. 

Following the coarse-to-fine framework, the search range is estimated using confidence output from the confidence estimation module, calculated as:
\vspace{-7pt}
\begin{equation}
\begin{aligned}
    \sigma ^i&=\sqrt{\sum_{d^i}{(}d-\hat{d}^i)^2\times softmax(-c_{d}^{i})}, 
    \\
    \hat{d}^i&=\sum_{d^i}^{}{d}\times softmax(-c_{d}^{i}),
\end{aligned}
\end{equation}
\vspace{-10pt}

where $c$ represents the predicted cost tensor. Then, the disparity range of the next stage can be computed based on the standard deviation $\sigma^i$ as:
\vspace{-7pt}
\begin{equation}
\begin{aligned}
d_{max}^{i-1}&=\delta (\hat{d}^i+(s^i+1)\sigma ^i+\varepsilon ^i), \\d_{min}^{i-1}&=\delta (\hat{d}^i-(s^i+1)\sigma ^i-\varepsilon ^i), 
\end{aligned}
\end{equation}
\vspace{-12pt}

where $\delta$ denotes the bilinear upsampling operation. $s^i, \varepsilon^i$ are two learnable normalization factors with initial value 0. 

The dynamic estimation of the disparity search range based on confidence allows network to reduce computational costs while ensuring accuracy in challenging initial regions. Besides, the confidence output of the core model will be utilized in CBEM module. The multi-scale disparities are optimized by unsupervised loss marked with orange box in Fig.\ref{fig:frame} and will be discussed in detail in \ref{unsuppart}.
\vspace{-17pt}
\subsection{CBEM: Bridging confidence with disparity error}
\begin{figure}[htb]

    \centering
    \begin{subfigure}{0.3\columnwidth}
        \includegraphics[width=1\columnwidth]{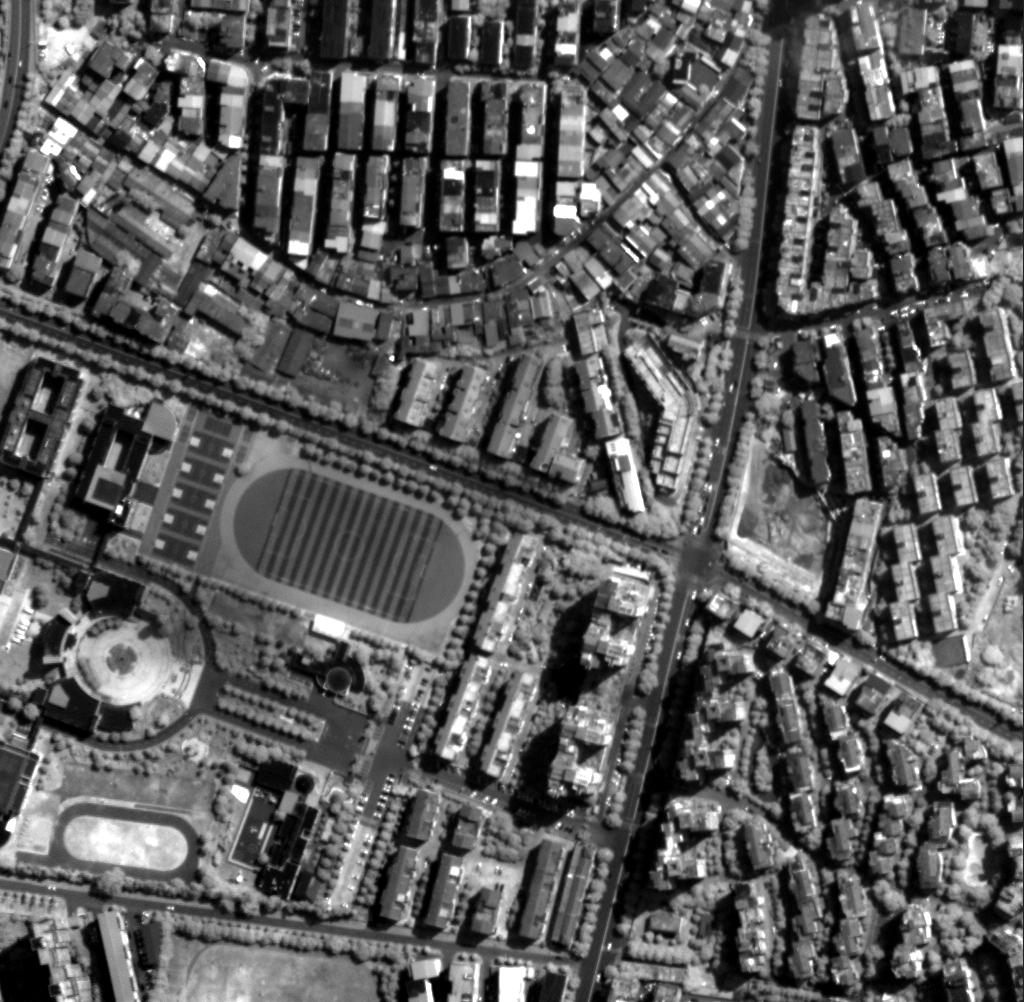}
    \end{subfigure}
    \begin{subfigure}{0.3\columnwidth}
        \includegraphics[width=1\columnwidth]{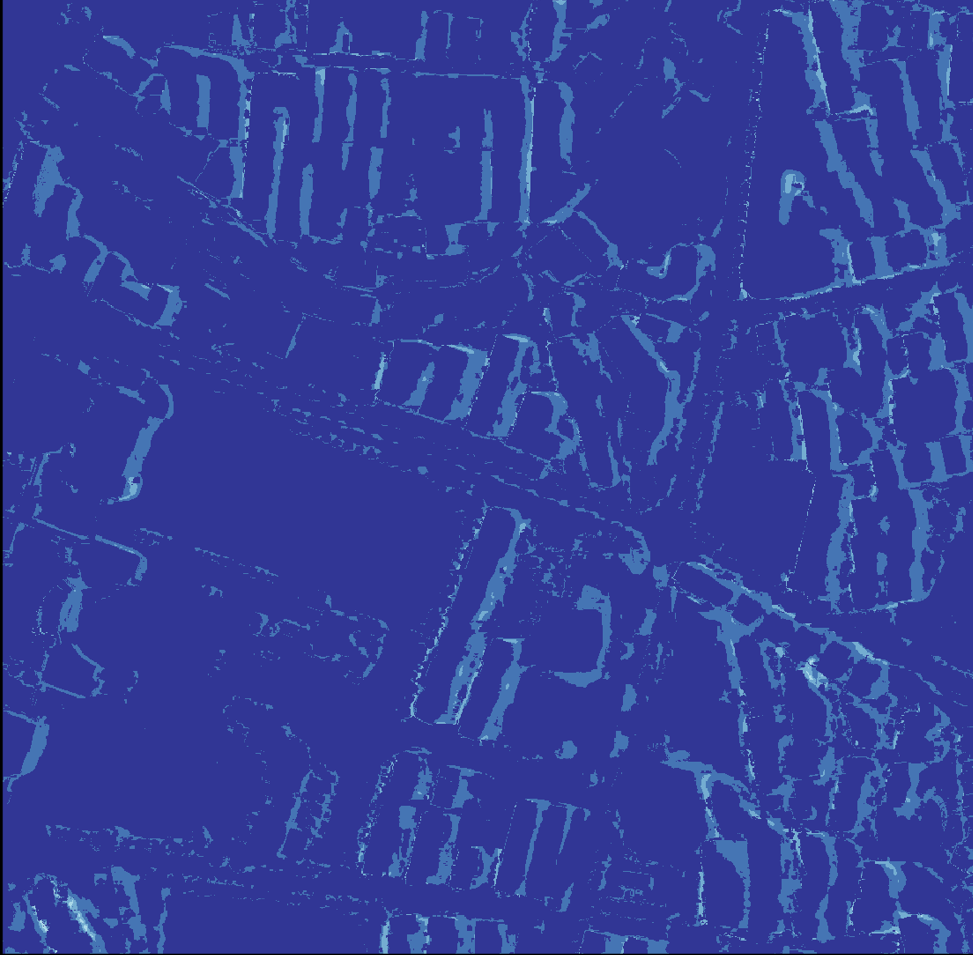}
    \end{subfigure}
    \begin{subfigure}{0.3\columnwidth}
        \includegraphics[width=1\columnwidth]{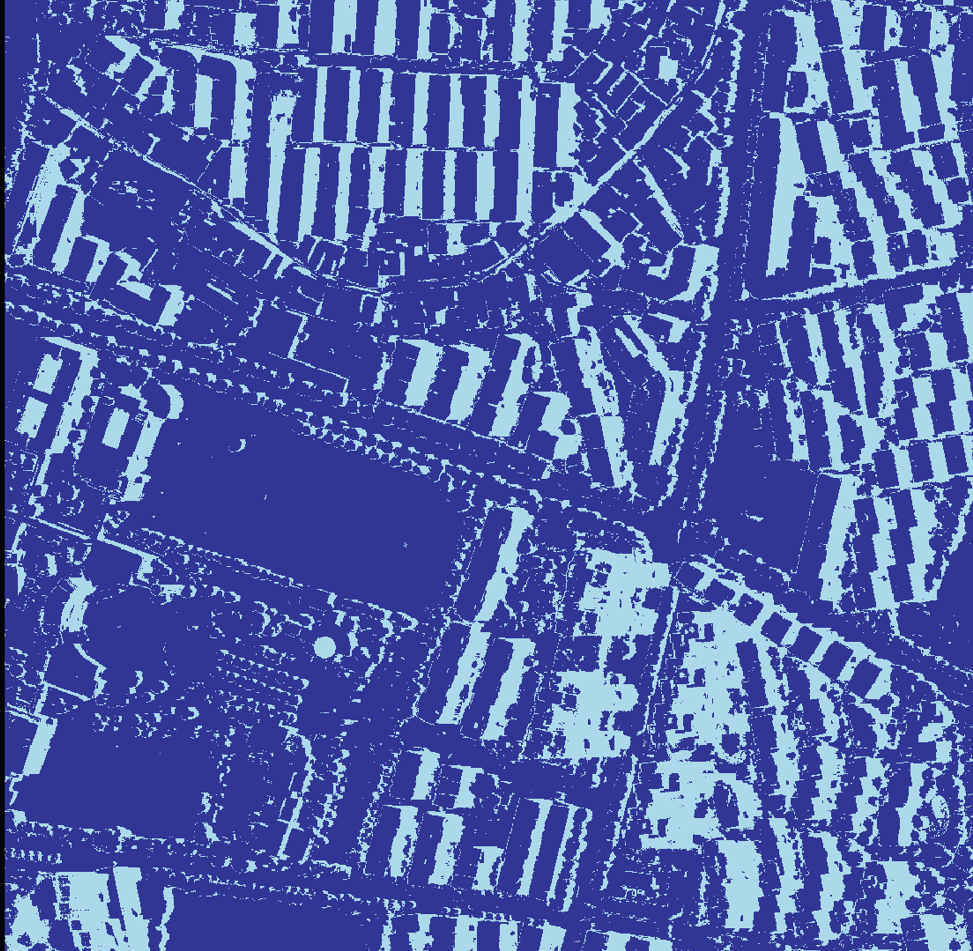}
    \end{subfigure}
    \begin{subfigure}{0.3\columnwidth}
        \includegraphics[width=1\columnwidth]{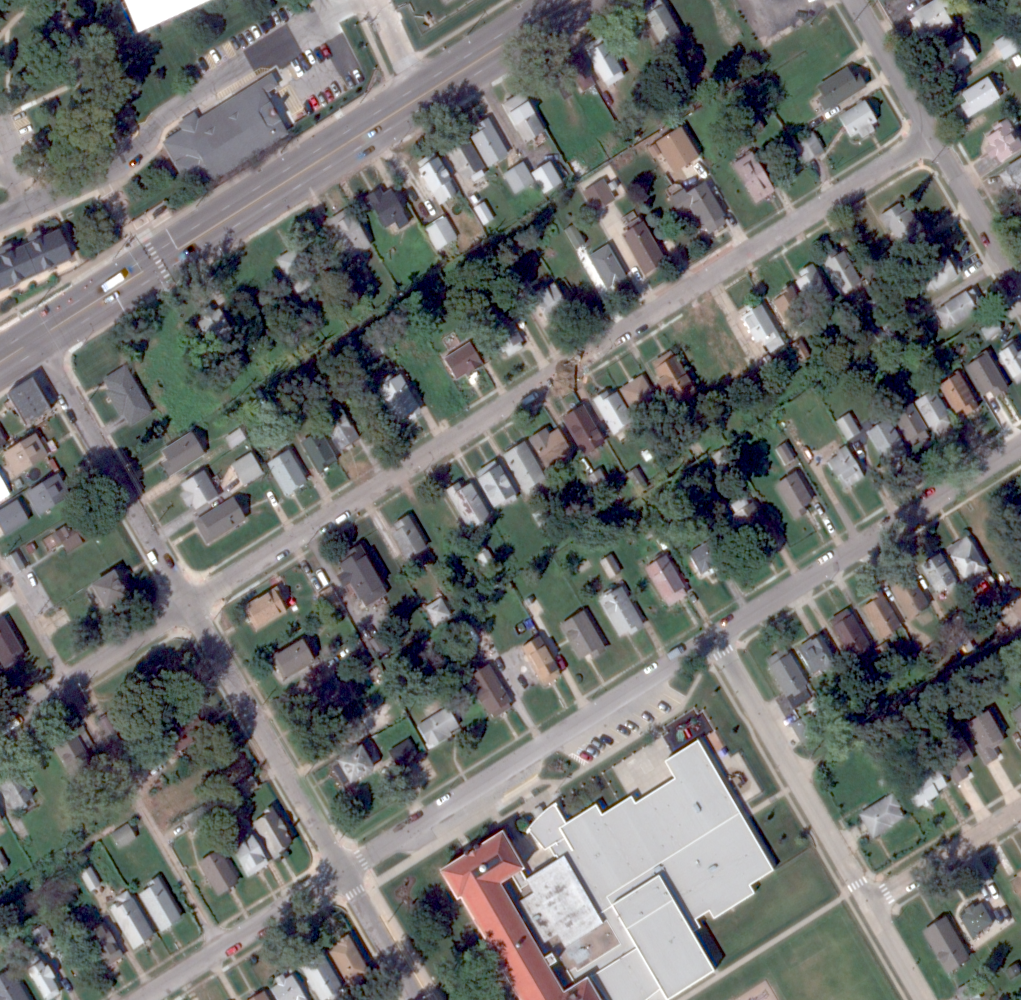}
        \caption{left image}
    \end{subfigure}
    \begin{subfigure}{0.3\columnwidth}
        \includegraphics[width=1\columnwidth]{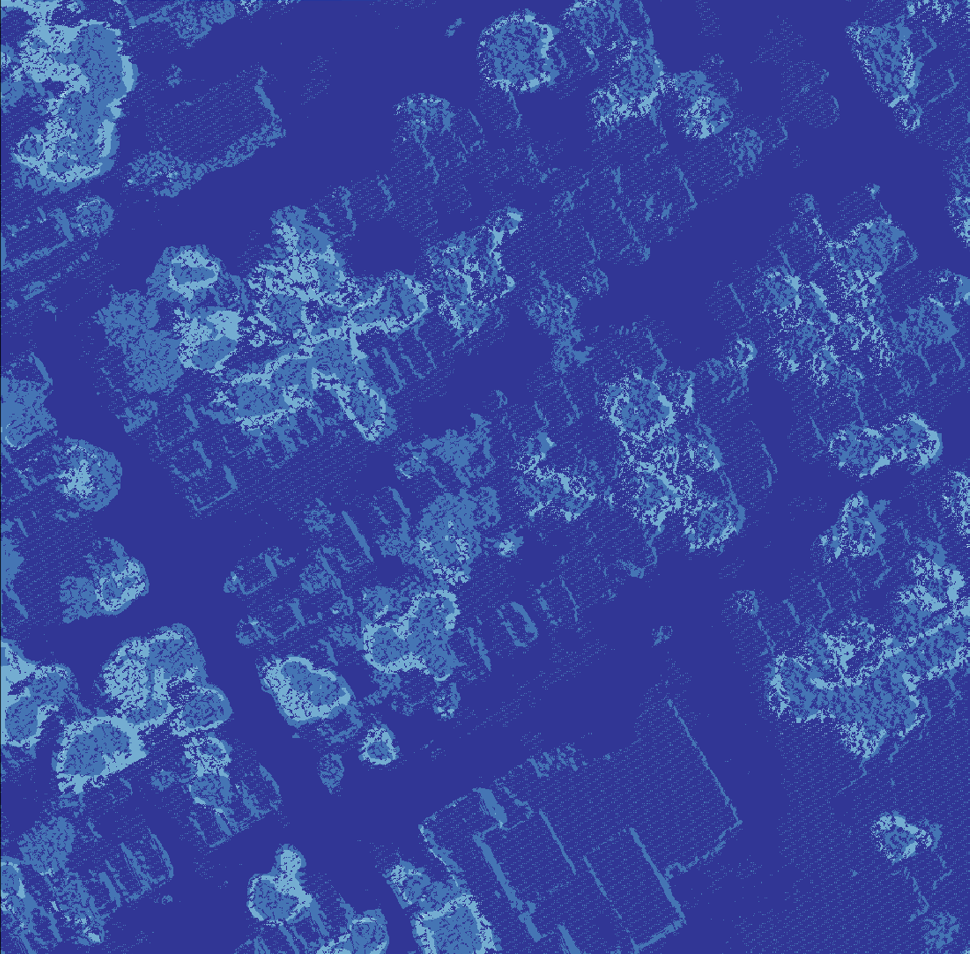}
        \caption{uncertainty score}    
    \end{subfigure}
    \begin{subfigure}{0.3\columnwidth}
    \includegraphics[width=1\columnwidth]{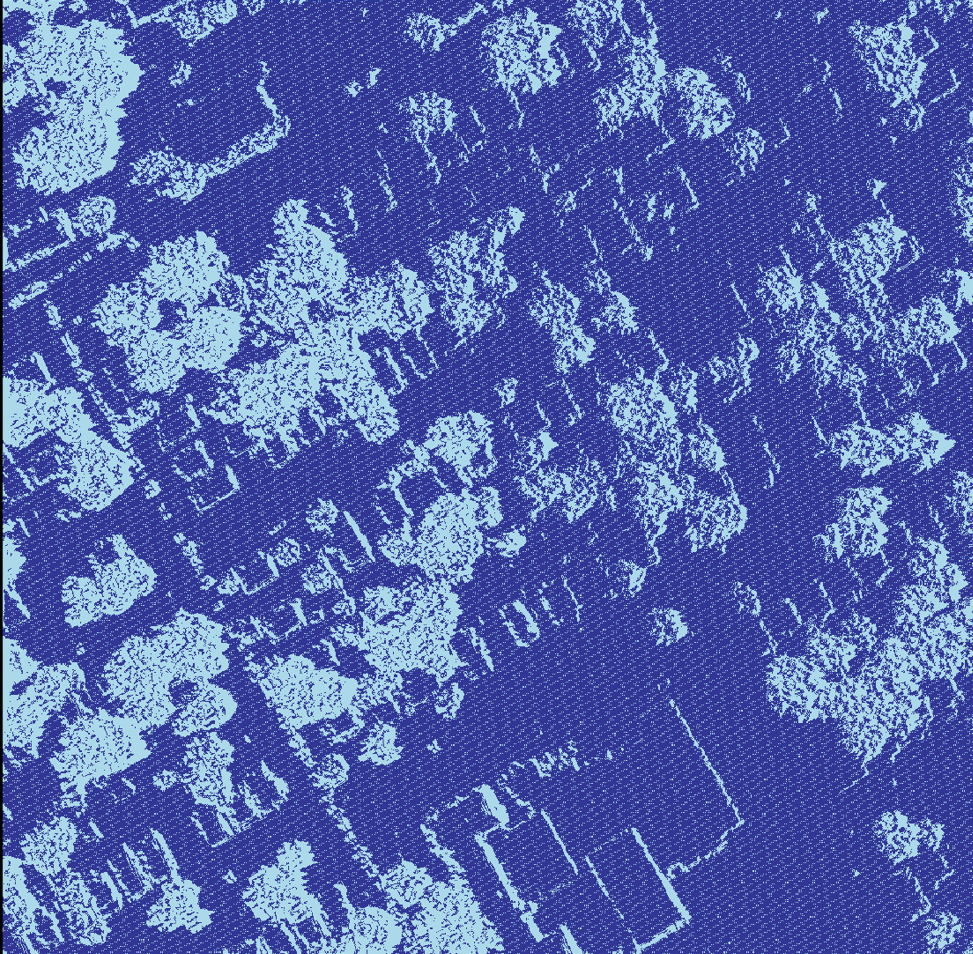}  
    \caption{disparity error}
    \end{subfigure}
\vspace{-7pt}
    \caption{CBEM performance on: Top: WHU-Stereo; Bottom: US3D. Warmer colors indicate higher levels of uncertainty or larger disparity errors. The uncertainty score is strongly correlated to disparity error for both WHU-Stereo and US3D.}
    \label{fig:conf}
\vspace{-10pt}
\end{figure}

\vspace{-7pt}
Motivated by the need to create a module that serves as an intermediary between confidence and true disparity error in order to reduce errors in disparity estimation, we developed the Confidence-Based Error Prediction Module(CBEM). It’s not feasible to predict accurate disparity error, otherwise it will be possible to reduce error thoroughly. Thus, we train a simple yet efficient hourglass module to output uncertainty score that aligns with the disparity error distribution. Unlike previous approaches that utilized the disparity to predict uncertainty scores, we used the confidence output from the core model as the input for our module. Although the core model is trained following unsupervised manner, we train CBEM with a few labelled data from WHU-Stereo for only one epoch to gain error prediction ability. Fig.~\ref{fig:conf} demonstrates the CBEM's stable error prediction capability when directly evaluated on the target US3D dataset. Notably, the CBEM exhibits strong cross-domain generalization ability, as it is less affected by the specific characteristics of original domain, owing to its reliance on confidence rather than disparity as input.

Here, we obtain the final predicted error by jointly encoding LR error and initial uncertainty. LR error is the difference between left image and reconstructed left image using right image and left disparity, as Eq.~\ref{eq:LRerror}, where $\mathcal{W}$ represents projection operation.The initial uncertainty is generated by core model, thus restricted by its performance, while LR error is not affected by the model. LR error depends on the projection error of left and right images, influenced by local noise and intensity inconsistency of images, while initial uncertainty does not have this shortcoming. Therefore, these two types of information complement each other well. We firstly encode LR error and initial uncertainty separately by convolution operation. After concatenating them in channel dimension, use two cascaded hourglass modules to learn the residuals of predicted disparity error.  
\begin{small}
\begin{equation}
	E_{LR} = \mathcal{W}(I_r, d_l)- I_l
	\label{eq:LRerror}
\end{equation}
\end{small}
\vspace{-7pt}

Binary Cross-Entropy loss is adopted to train the CBEM module. Ground truth disparity error above threshold(1.0) is noted as 1, otherwise 0. Data used to train CBEM comes from other datasets with ground truth. Due to its strong cross-domain generalization capability, CBEM still functions well in unsupervised dataset. The formula is as follows.
\vspace{-7pt}
\begin{equation}
\begin{aligned}
GT_{error}&=\left\{
\begin{array}{rcl}
1.0       &      & {|GT_{disparity}-\hat{d}|>      1.0}\\
0.0     &      & {otherwise}
\end{array} \right. \\
L_{BCE}&=BCE(uscore,GT_{error})
\end{aligned}
\label{eq:gt}
\end{equation}
\vspace{-7pt}
\subsection{Unsupervised Loss}
\vspace{-7pt}
\label{unsuppart}
For the first stage of training, we optimize the core model using unsupervised loss of three scales. The loss function is the same for each scale, denoted $L_S$. The full loss is the sum of individual scale losses, i.e., $\mathcal{L}=\sum^{}_{s}\mathcal{L}_s$.
The $\mathcal{L}_S$ combines three terms, given as:
\vspace{-7pt}
\begin{equation}
\mathcal{L}_S=\lambda_{ap}\mathcal{L}_{ap}+\lambda_{census}\mathcal{L}_{census}+\lambda_{sm}\mathcal{L}_{sm}
\vspace{-3pt}
\end{equation}
 \vspace{-7pt}
\begin{small}
\begin{equation}
	\begin{aligned}
	\mathcal{L}_{ap} & = \frac{1}{N} \sum_{i,j}^{}\alpha \frac{1-SSIM(I_{i,j}^{l} \odot (1 - O_{i,j}^{l}),\tilde{I}_{i,j}^{l} \odot (1 - O_{i,j}^{l}) ) }{2}  \\
	&+ (1-\alpha )\left \| I_{i,j}^{l} \odot (1 - O_{i,j}^{l}),\tilde{I}_{i,j}^{l} \odot (1 - O_{i,j}^{l})  \right  \|   
	\end{aligned}
	\label{eq:re}
\end{equation}
\end{small}
\vspace{-7pt}

$L_{ap}$ promotes consistency between left/right image and its reconstructed image in non-occluded areas as Eq.~\ref{eq:re}, while we use a simple formulation to detect occluded areas by distance of forward-backward disparities, defined as Eq.~\ref{eq:occ}:
\begin{equation}
    \begin{aligned}
    |d_i^f(j)+d_i^b(d^f_i(j))|^2<\tau_i 
    \end{aligned}
    \label{eq:occ}
\end{equation}
where $d^f(j)$ represents the forward disparity based on left image,  $d^b(j)$  represents backward disparity  based on right image,  $O_{i,j}^{l}$ is the predicted occlusion map. $\odot$ denotes element-wise multiplication, $\alpha$ is set to 0.85. $\tau_i=[5,2,1]$ according to the general pattern of VHR images.

$\mathcal{L}_{census}$ is also adopted to reduce reconstruction errors while it has better robustness to various brightness and shadow disturbances in VHR remote sensing images. We utilize Census transform($\tau$) with a patch size of 7 pixels and Charbonnier penalty function($\rho$) to implement $L_{census}$, Hamming Distance denoted as $\Gamma$. The formulation is as follow:
\begin{equation}
	\mathcal{L}_{census}  = \frac{1}{N} \sum_{i,j}^{} \rho[\Gamma(\tau(I_{i,j}^{l}) ,\tau(\tilde{I}_{i,j}^{l}))] \odot (1 - O_{i,j}^{l}) ) 
\end{equation}

$\mathcal{L}_{sm}$ is utilized to smooth disparity with $L_2$ penalty on disparity gradients $\partial{d}$ and image gradients $\partial{I}$, denoted as:
\begin{equation}
    \mathcal{L}_{sm} = |\partial{_x}d_L|e^{-|\partial{_x}I_L|}+|\partial{_y}d_L|e^{-|\partial{_y}I_L|}  
\end{equation}

For the second stage of training, CBEM is fixed to generate a reliable uncertainty score map. As CBEM has gained ability to generate uncertainty score correlated with disparity error, by using fine-scale disparity to supervise coarse-scale disparity in reliable regions, we aim to further refine disparity. Here, we use a mask $M=[uscore<t]$ to exclude poor-performing regions, only focusing on those areas that are good enough. $\mathcal{L}_{self-sup}$ is calculated as follows:
\begin{equation}
    \mathcal{L}_{self-sup}=\frac{1}{N}\sum_{i,j}[smooth_{L1}(d_{i,L}- d_{i,j})\odot M_{i}]
\end{equation}

\vspace{-17pt}
\section{EXPERIMENTAL RESULTS}
\vspace{-7pt}
\subsection{Dataset}
The \textbf{SceneFlow dataset}~\cite{sceneflow} in the finalpass version is used to pretrain our core network, which includes 35,454 positive training and 4370 test images with a resolution of $960\times540$, so as the negative ones. \textbf{US3D track-2 dataset}~\cite{us3d} of the 2019 Data Fusion Contest is used to evaluate our network, which contains VHR images collected by WorldView-3 between 2014 and 2016 over Jacksonville (JAX) and Omaha (OMA) in the United States. 1712 RGB stereo pairs of JAX are used for training while the remaining 427 pairs are for validation. All 2153 pairs of OMA are used for testing. We also use the with-ground-truth subset of \textbf{WHU-Stereo dataset}~\cite{whu} to validate the cross-domain generalization ability, which contains 1757 GaoFen-7 panchromatic image pairs over Kunming, Qichun, Yingde, Shaoguan, Wuhan and Hengyang in 2020. For training CBEM, we also borrow this dataset to establish relationship between confidence and disparity errors.
\vspace{-12pt}
\subsection{Implementation Details}
\begin{figure}[tb]
\vspace{-7pt}
\centering
	\includegraphics[width=0.8\columnwidth]{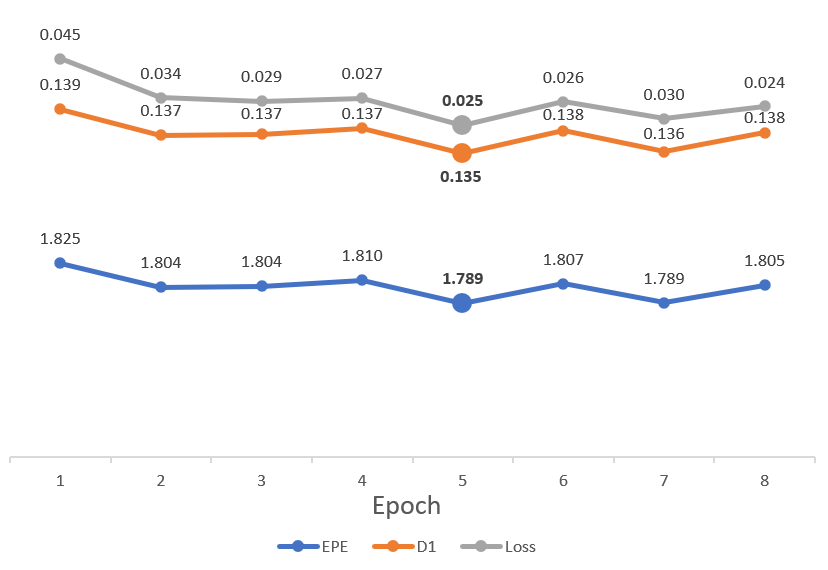}

\caption{The validation loss curve.}
\label{fig:curve}
\vspace{-10pt}
\end{figure}

\vspace{-7pt}
Our network is implemented with PyTorch 1.9.0 and is trained with Adam($\beta_1=0.9,\beta_2=0.999$ ) . All input stereo images are randomly cropped to $512\times512$ for training while $1024\times1024$ original size is used for testing. We first pretrain our core model on SceneFlow dataset from scratch for 20 epochs with learning rate 0.001 and batch size 8. Then core model is finetuned on US3D dataset in unsupervised manner for 60 epochs with learning rate set to 0.0001. The disparity range is set to $[-128, 128]$. After that, we fix the core model and train CBEM using WHU-Stereo , which only takes 1 epoch with learning rate 0.001 to achieve good error prediction ability. Last step is to fix CBEM and finetune core model with learning rate 1e-5 for 6 epochs. All experiments are conducted on two NVIDIA Titan-RTX GPUs. To avoid overfitting, we stop training as soon as  $\mathcal{L}_{self-sup}$ increases, even though it may drop later. As shown in Fig.~\ref{fig:curve} on validation set, we plot EPE and D1 curves to demonstrate the need for an immediate stop, Epoch 5 is the best epoch to be adopted.
\vspace{-17pt}
\subsection{Performance}
We conducted a comparative analysis between supervised models, namely PSMNet~\cite{psmnet} and CACN~\cite{CACN}, as well as unsupervised methods including SGBM+WLS~\cite{SGM} and T.Igeta's model~\cite{igeta}, against the proposed method. The evaluation results on US3D dataset are presented in Table.~\ref{table:us3d} using EPE and D1 as metrics. It is evident that our model outperforms other unsupervised models and gains competitive results compared to supervised models. When comparing with core model, our method improves 11.7 \% for EPE and 16.6 \% for D1, which proves validity of CBEM.

\begin{figure}[tb]
    \centering
    \begin{subfigure}{0.24\columnwidth}
        \includegraphics[width=1\columnwidth]{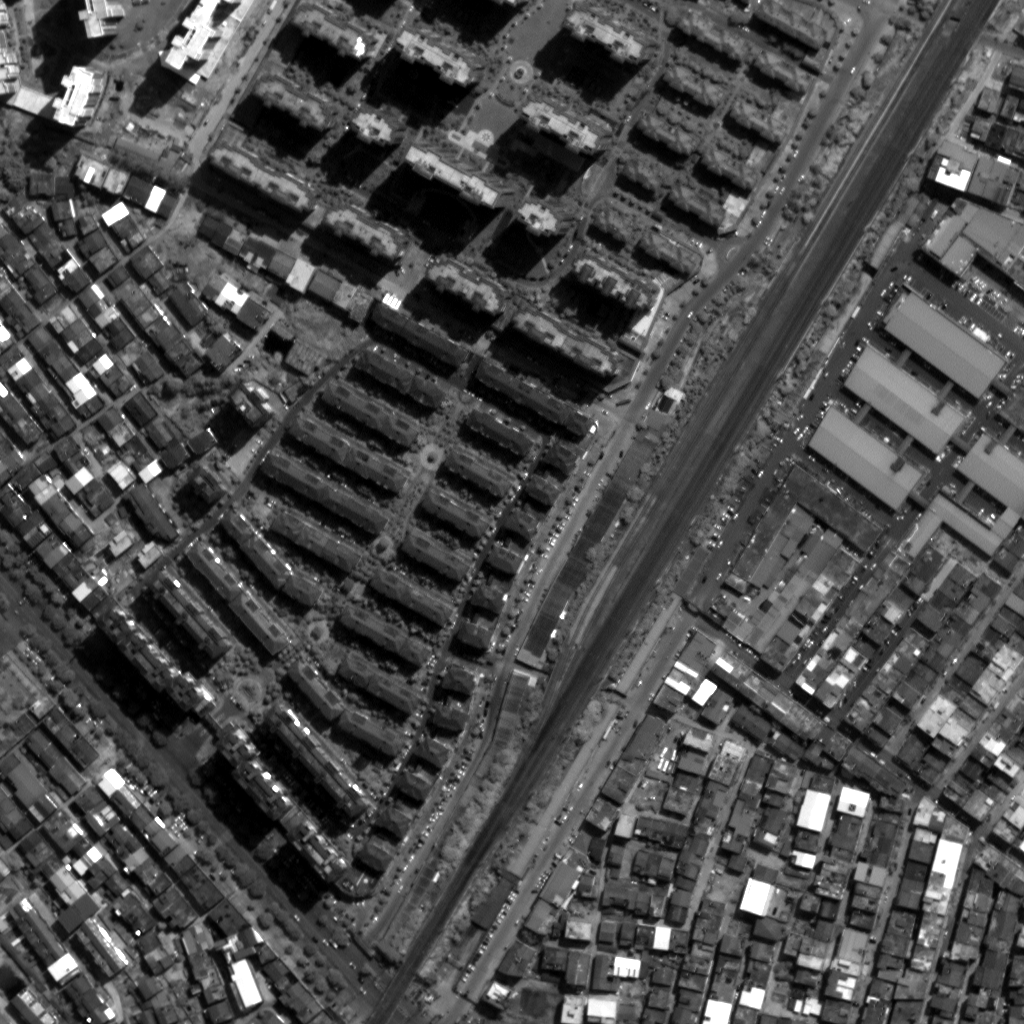}
    \end{subfigure}
    \begin{subfigure}{0.24\columnwidth}
        \includegraphics[width=1\columnwidth]{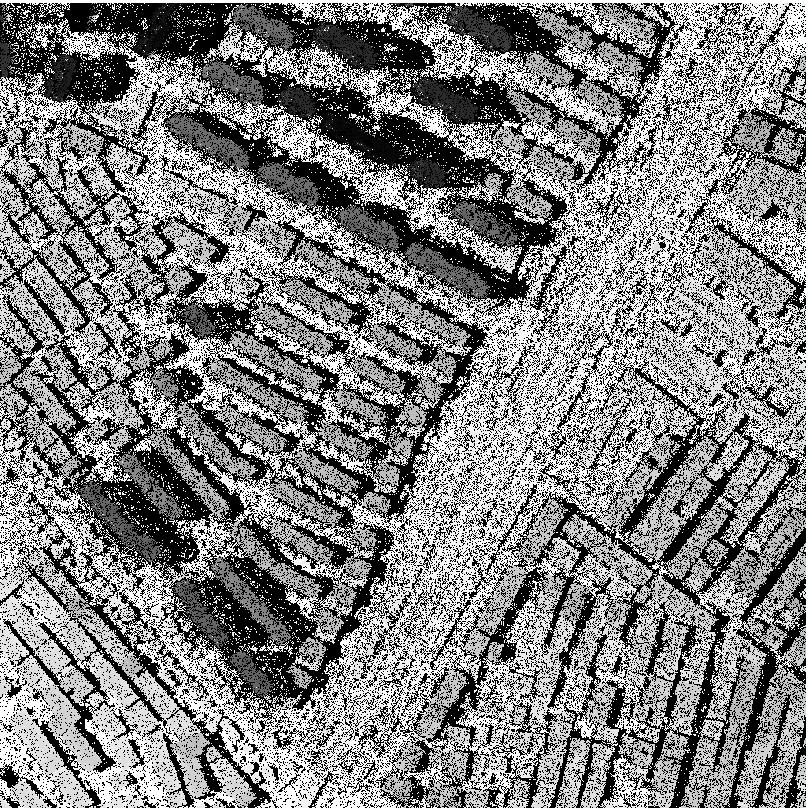}
    \end{subfigure}
    \begin{subfigure}{0.24\columnwidth}
        \includegraphics[width=1\columnwidth]{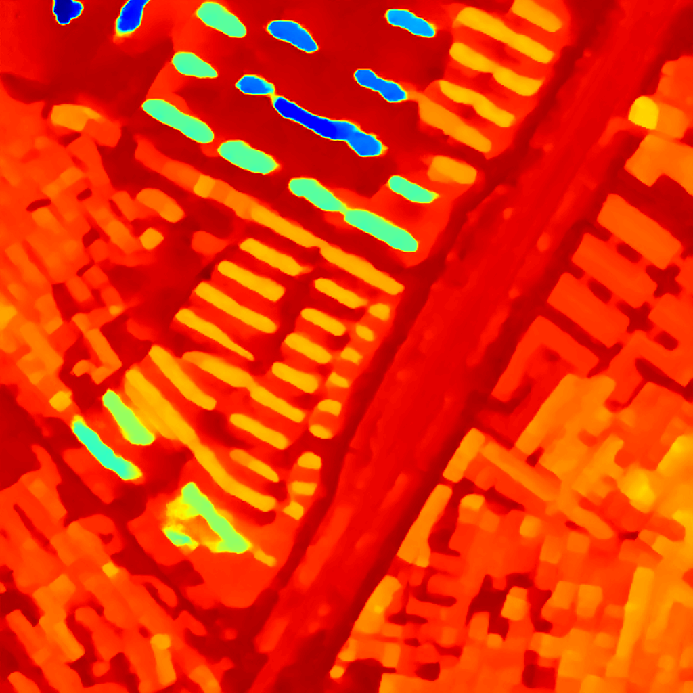}
    \end{subfigure}
    \begin{subfigure}{0.24\columnwidth}
        \includegraphics[width=1\columnwidth]{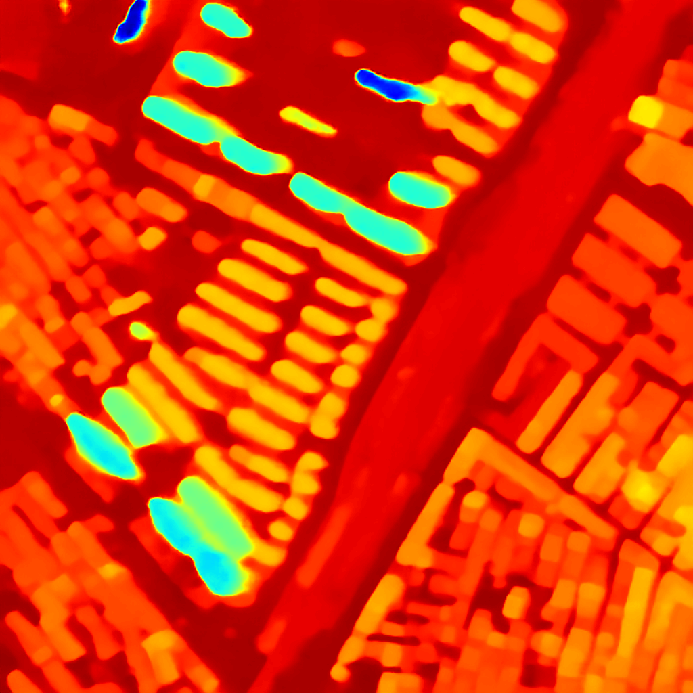}
    \end{subfigure}
    \begin{subfigure}{0.24\columnwidth}
        \includegraphics[width=1\columnwidth]{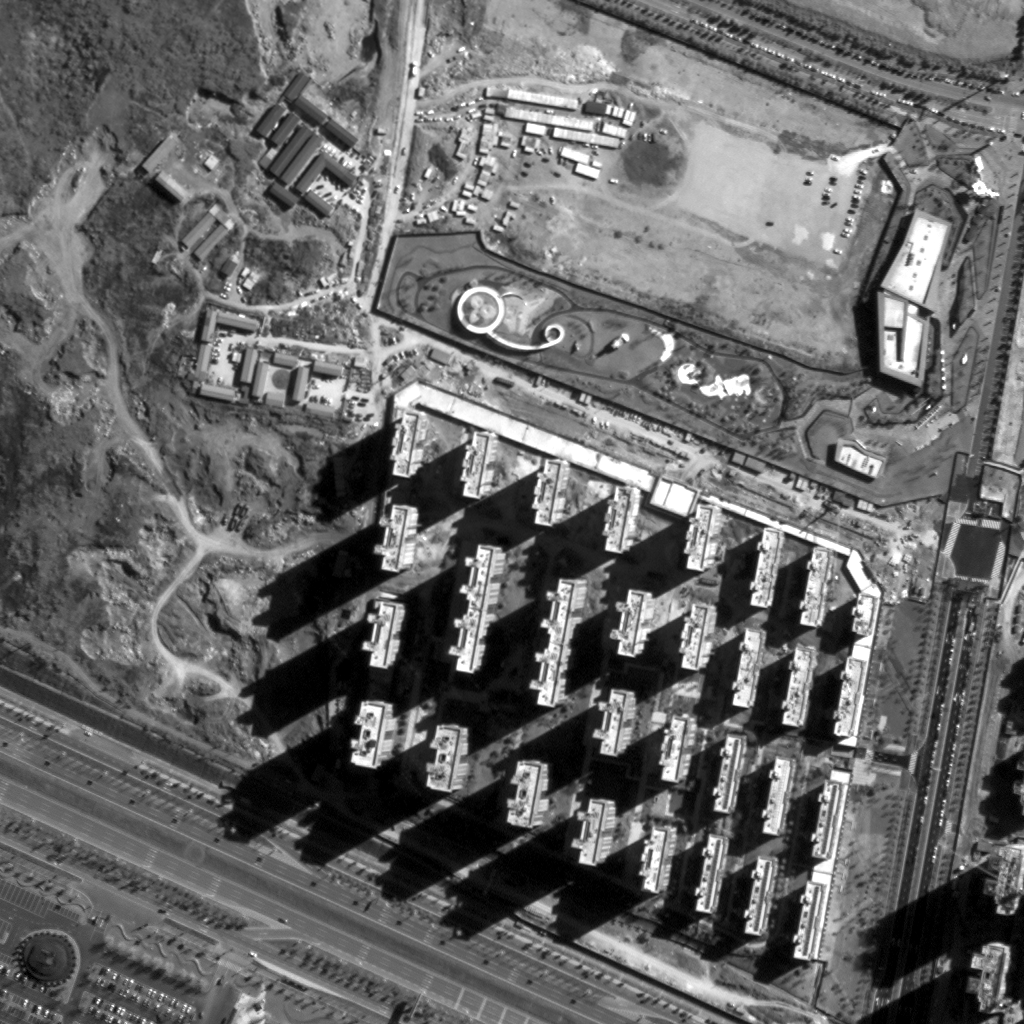}
    \end{subfigure}
    \begin{subfigure}{0.24\columnwidth}
        \includegraphics[width=1\columnwidth]{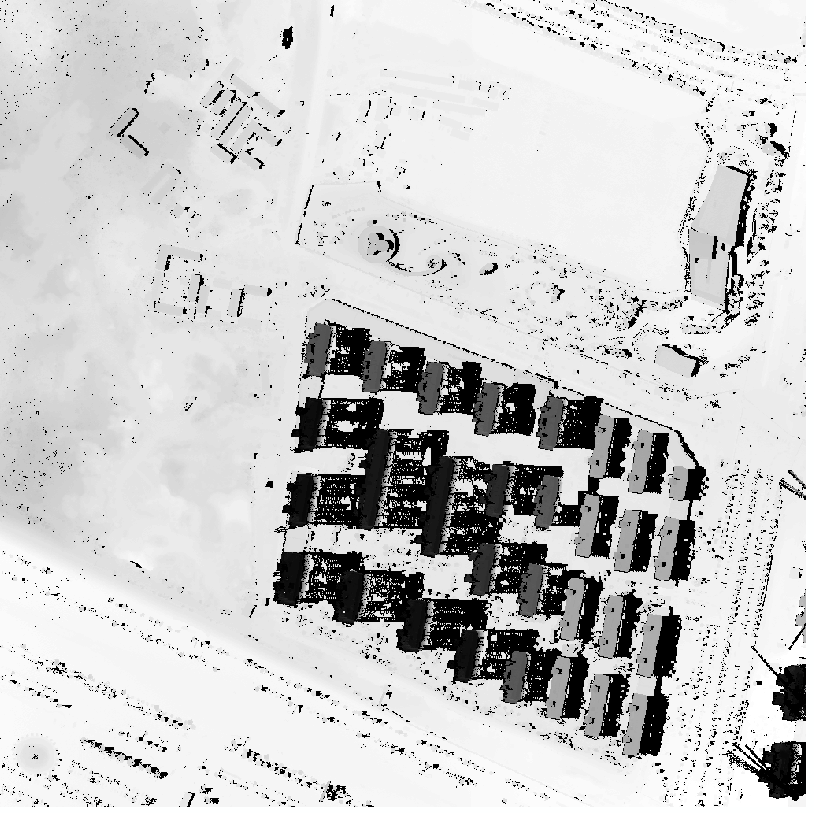}
    \end{subfigure}
    \begin{subfigure}{0.24\columnwidth}
        \includegraphics[width=1\columnwidth]{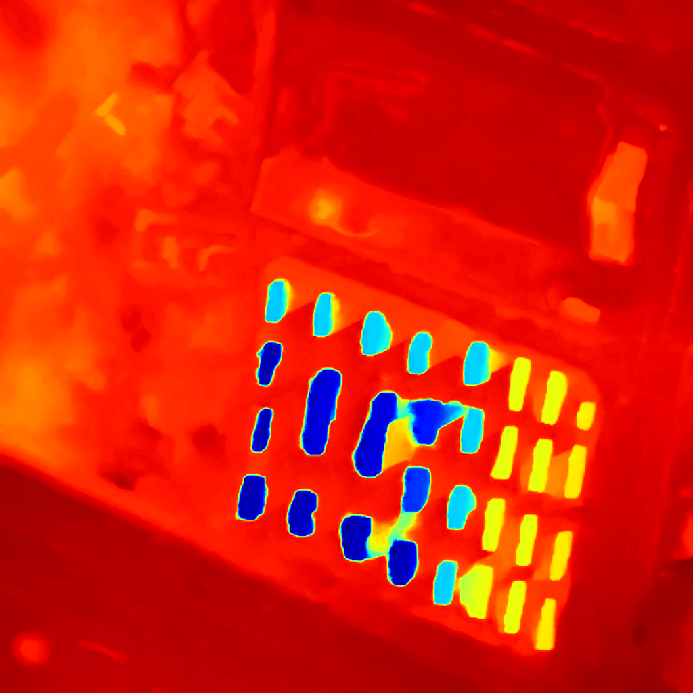}
    \end{subfigure}
    \begin{subfigure}{0.24\columnwidth}
        \includegraphics[width=1\columnwidth]{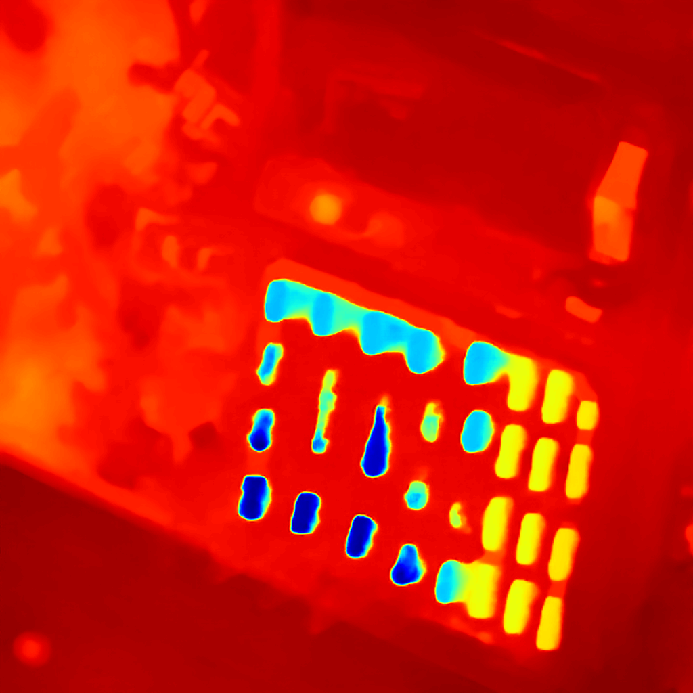}
    \end{subfigure}
    \begin{subfigure}{0.24\columnwidth}
        \includegraphics[width=1\columnwidth]{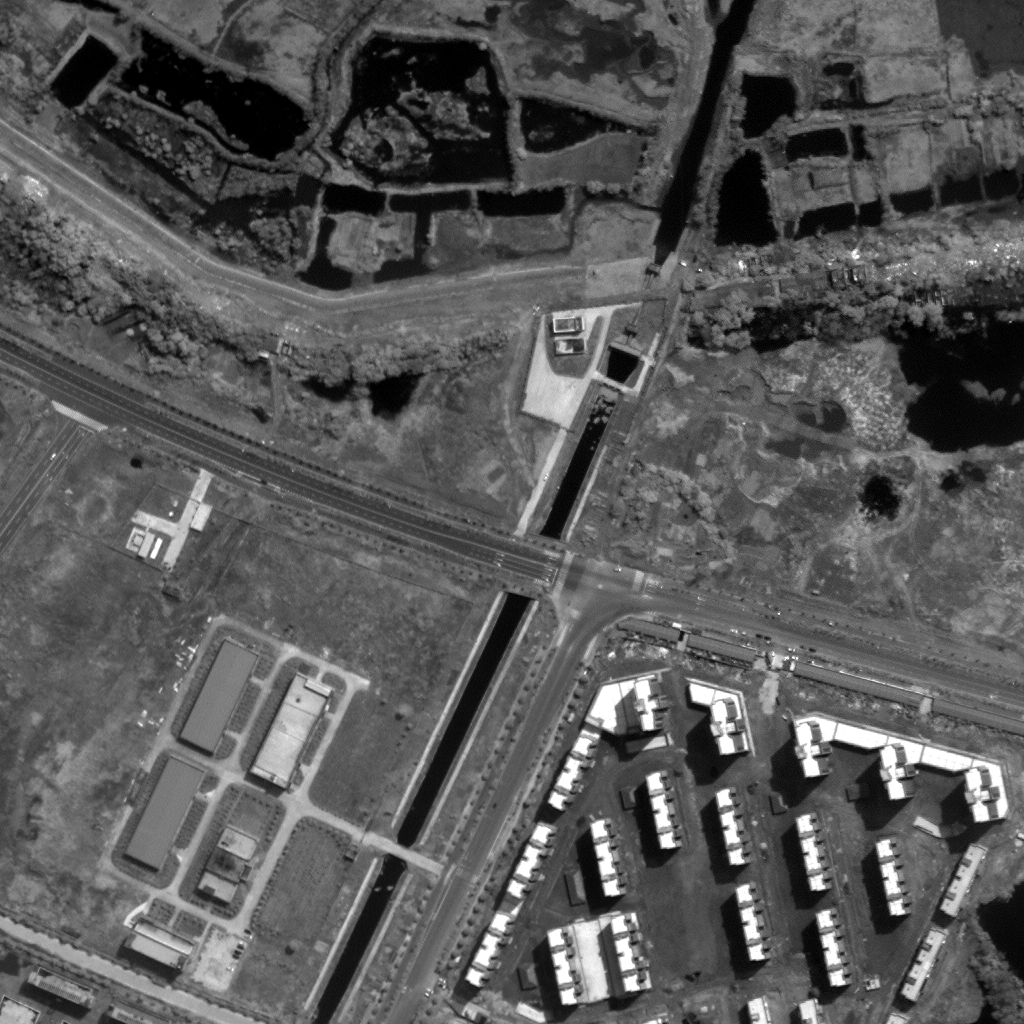}
    \end{subfigure}
    \begin{subfigure}{0.24\columnwidth}
        \includegraphics[width=1\columnwidth]{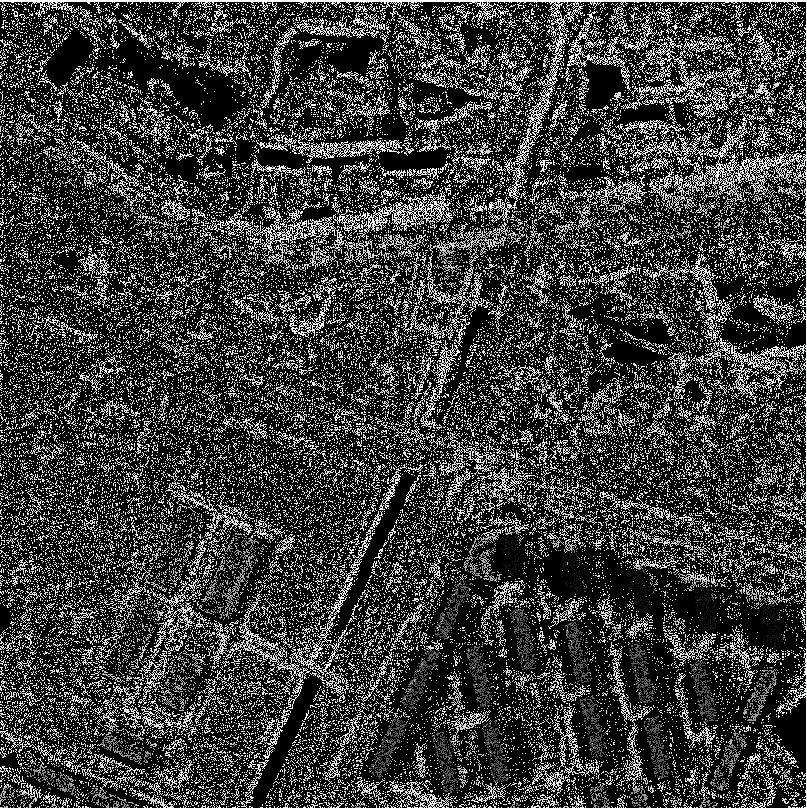}
    \end{subfigure}
    \begin{subfigure}{0.24\columnwidth}
        \includegraphics[width=1\columnwidth]{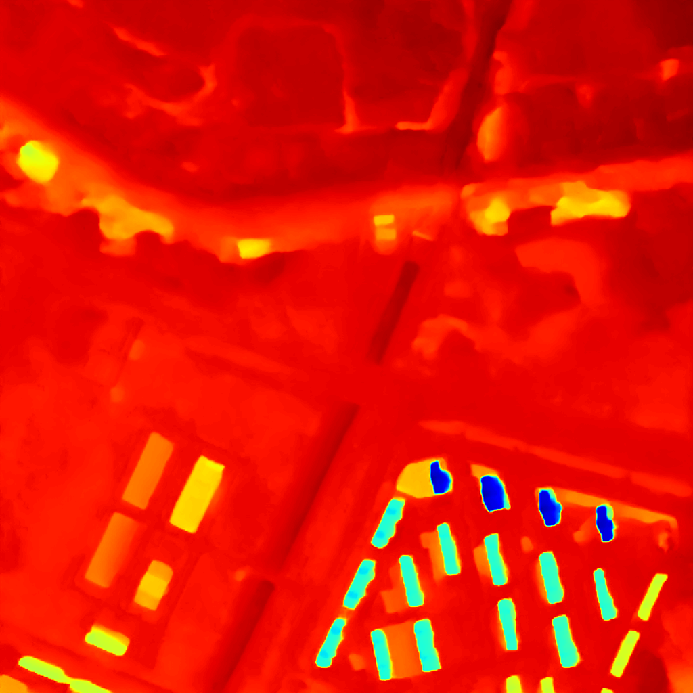}
    \end{subfigure}
    \begin{subfigure}{0.24\columnwidth}
        \includegraphics[width=1\columnwidth]{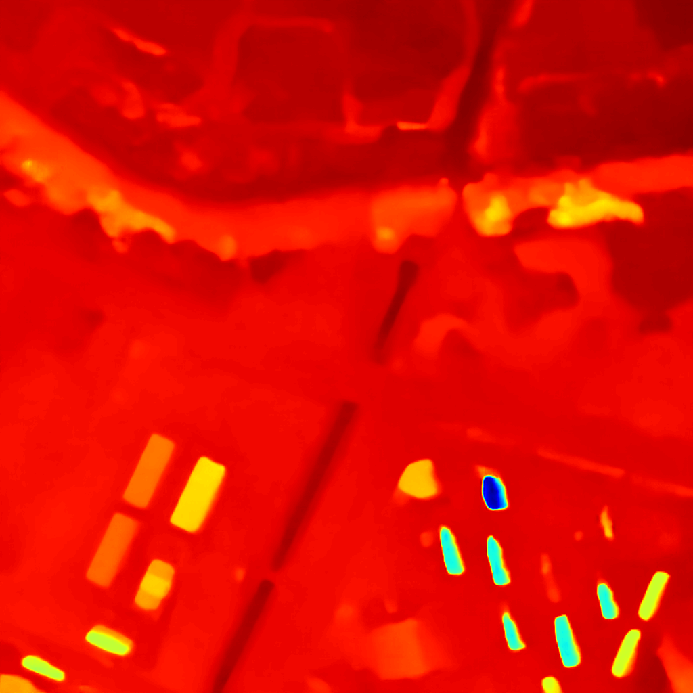}
    \end{subfigure}
    
    \caption{Disparity estimation on WHU-Stereo. From left to right: left image, GT, proposed, CACN~\cite{CACN}.}
\vspace{-12pt}
    \label{fig:whu}
\end{figure}

The core model trained on US3D dataset is directly used to test the WHU-Stereo dataset, revealing the superior cross-domain generalization ability of our proposed unsupervised model, according to Table.~\ref{table:whu}. As shown in Fig.~\ref{fig:whu}, the proposed method generates disparity with clearer edges of buildings due to smoothing function of $\mathcal{L}_{sm}$. Furthermore, the proposed method is less prone to missing fuzzy buildings with shadows, owing to the robustness to photometric inconsistency provided by $\mathcal{L}_{census}$. In the disparities generated by CACN model, building edges usually appear dilated, resulting in an overestimation of the building's size. We have mitigated this issue by incorporating the CBEM module, which reduces predicted disparity error and effectively preserves the precise delineation of building edges.

\vspace{-7pt}
\begin{table}[h]
\caption{Performance of comparison methods on US3D. }    
\centering
\scalebox{0.8}{
\begin{tabular}{ccccc}
\hline
Method &with GT &Deep Learning &EPE &D1(\%) \\
\hline
PSMNet~\cite{psmnet} &yes &yes &1.48 &8.53 \\
CACN~\cite{CACN} &yes &yes &1.47 & 8.23 \\
\hline
SGBM+WLS~\cite{sgbm} &no &no &2.56 &15.9 \\
T.Igeta et al.~\cite{igeta} &no &yes &2.12 &15.3 \\
core model &no &yes &\bf{1.96} &\bf{14.5} \\
Proposed &no &yes &\bf{1.73} &\bf{12.1} \\
\hline
\label{table:us3d}
\end{tabular}}
\vspace{-10pt}
\end{table}
\vspace{-7pt}
\begin{table}[!h]
\caption{Performance on WHU-Stereo}
\centering
\scalebox{0.85}{
\begin{tabular}{cccc}
\hline
Method &with GT &EPE &D1(\%) \\
\hline
CACN &yes  &3.47 &33.5 \\ 
core model &no  &\bf{3.11} &\bf{31.5} \\
\hline
\label{table:whu}
\end{tabular}}
\vspace{-10pt}
\end{table}
\vspace{-7pt}
\section{Conclusions}
\vspace{-7pt}
In this study, we propose a novel unsupervised stereo matching network based on error prediction for VHR images, which bridges confidence with disparity error, demonstrating robust cross-domain generalizability. For future work, we aim to broaden the scope of the model’s applicability.
\vspace{-7pt}
\section{ACKNOWLEDGEMENT}
\vspace{-7pt}
The authors would like to thank the Johns Hopkins University Applied Physics Laboratory and the IARPA for providing the data used in this study, and the IEEE GRSS Image Analysis and Data Fusion Technical Committee for organizing the Data Fusion Contest. The authors extend their gratitude to Li et al. for supplying the WHU-Stereo dataset utilized in this study. This work was supported by Key Research Program of Frontier Sciences, Chinese Academy of Sciences, under Grant ZDBS-LY-JSC036.

\bibliographystyle{IEEEbib}
\bibliography{strings,refs}

\end{document}